\documentclass[letterpaper, 10 pt, conference]{ieeeconf}  

\usepackage{amsmath,amssymb,amsfonts,float,graphicx,bm,psfrag,amsmath}
\usepackage{algorithm, algorithmic,CJK,color,color,geometry,url}
\usepackage{graphicx,cite,bm,psfrag,amsmath}
\usepackage{booktabs}
\usepackage{balance}

\newcommand{\tabincell}[2]{\begin{tabular}{@{}#1@{}}#2\end{tabular}}

\IEEEoverridecommandlockouts                              

\begin{document}
\title{\bf
TJ-FlyingFish: Design and Implementation of an Aerial-Aquatic Quadrotor with Tiltable Propulsion Units
}

\author{Xuchen Liu$^{1,2}$, Minghao Dou$^{1,2}$, Dongyue Huang$^{1,2}$, Biao Wang$^{3,4}$, Jinqiang Cui$^{4}$, \\\hspace{1.2cm}Qinyuan Ren$^{5,4}$, Lihua Dou$^{6}$, Zhi Gao$^{7}$, Jie Chen$^{1}$ and Ben M. Chen$^{2}$
\thanks{$^{1}$Shanghai Research Institute for Intelligent Autonomous Systems, Tongji University, Shanghai, China. \protect\url{chenjie206@tongji.edu.cn}.}
\thanks{$^{2}$Department of Mechanical and Automation Engineering, the Chinese University of Hong Kong, Hong Kong, China. \protect\url{xcliu, mhdou, dyhuang@mae.cuhk.edu.hk}, \protect\url{bmchen@cuhk.edu.hk}.} 
\thanks{$^{3}$College of Automation Engineering, Nanjing University of Aeronautics and Astronautics, Nanjing, Jiangsu, China. \protect\url{wangbiao@nuaa.edu.cn}.}
\thanks{$^{4}$Peng Cheng Laboratory, Shenzhen, Guangdong, China. \protect\url{cuijq@pcl.ac.cn}.}
\thanks{$^{5}$College of Control Science and Engineering, Zhejiang University, Hangzhou, Zhejiang, China. \protect\url{latepat@gmail.com}.}
\thanks{$^{6}$School of Automation, Beijing Institute of Technology, Beijing, China. \protect\url{doulihua@bit.edu.cn}.}
\thanks{$^{7}$School of Remote Sensing and Information Engineering, Wuhan University, Wuhan, Hubei, China. \protect\url{gaozhinus@gmail.com}.}
}

\maketitle
\thispagestyle{empty}

\pagestyle{empty}

\begin{abstract}
Aerial-aquatic vehicles are capable to move in the two most dominant fluids, making them more promising for a wide range of applications. We propose a prototype with special designs for propulsion and thruster configuration to cope with the vast differences in the fluid properties of water and air. For propulsion, the operating range is switched for the different mediums by the dual-speed propulsion unit, providing sufficient thrust and also ensuring output efficiency. For thruster configuration, thrust vectoring is realized by the rotation of the propulsion unit around the mount arm, thus enhancing the underwater maneuverability. This paper presents a quadrotor prototype of this concept and the design details and realization in practice.
\end{abstract}

\section{INTRODUCTION}

Cross-domain collaboration has emerged as a hotspot in current research and application of unmanned system. However, coordination between aerial and aquatic domains is challenging, due to the restricted communication caused by the discontinuity of mediums. The aerial-aquatic vehicle is capable to shuttle in both 3-D fluid spaces, can act as an irreplaceable node linking the two unmanned systems, and enhancing sharing and fusion of information. Furthermore, the aerial-aquatic vehicle can be independently applied to scenarios where a single-medium one is not available, such as cross-domain surveys, remote sensing, disaster rescue, and so on.

Nonetheless, the significant differences in the environmental properties of the two medium present a design challenge for the vehicle. To begin with, designing a propulsion system highly depends on fluid density and viscosity since they affect output speed and torque. Secondly, the higher density and viscosity of water cause buoyancy and significant resistance to the vehicle, which leads to an different implementation to achieve static and dynamic equilibrium. The main goal in the air is to overcome gravity, while the major goal in the water is to achieve neutral buoyancy and overcome resistance to movement, resulting in different requirements for the thruster configuration of the vehicle.

\begin{figure}[tbp]
\centering
        \includegraphics[scale=0.25]{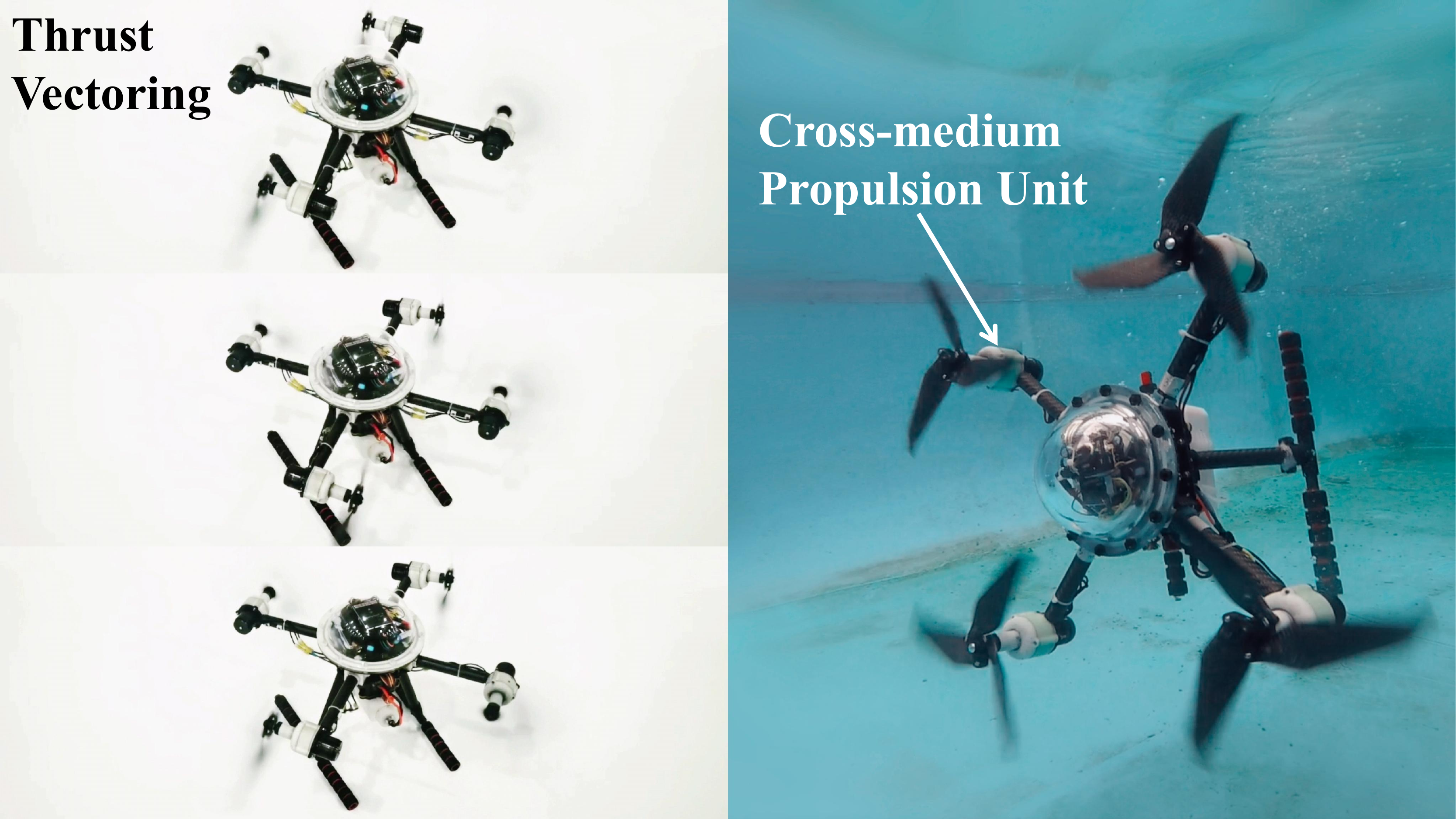}
\caption{Design essentials of the prototype.}\label{platform}
\end{figure}

Many functionalized prototypes have been demonstrated in recent years \cite{yu2021survey,zeng2022OE}, including the fixed-wing-based\cite{rockenbauer2021dipper}, the multirotor-based \cite{horn2019study,alzu2018loon,li2022science}, and the biologically inspired\cite{chen2017biologically}. For multirotor-based prototypes, while there exist platforms optimized for diverse fluid environments\cite{alzu2018loon}, the majority of them are just standard aerial hardware constructions with water resistance, which lack the same level of mobility in water as they do in the air. Another option is to splice submerged and airborne platforms together\cite{vyas2019modelling,bi2022nezha}, which increases the airborne load and hence impacts the aerial capacity.

In prior work, we proposed a morphable quadrotor \cite{yu2020icra} that allows for increased flexibility in thruster direction thanks to coupled symmetric thrust vectoring achieved through simple mechanical connections, which increases the mobility of the vehicle on the surface and underwater to a certain extent. However, due to the strong coupling in hydrodynamics and under-actuation of the prototype, the attitude controller is hard to design. In terms of the propulsion system, aerial motors and propellers are utilized to operate in both air and water by compromising air performance for underwater functionality \cite{yu2019iros}. 

The principal goal of this work is to facilitate the controller design and improve the performance of the propulsion system by the structural modification. To achieve this, an independent tilting mechanism is used to provide thrust vectoring instead of coupled symmetric one, allowing for more combinations of the thruster configuration to remove the limitation of the original coupling mechanism on controller design, and also making the prototype a potential over-actuated vehicle. Besides, for propulsion, the dual-speed gearbox driven by motor forward and reverse is equipped to provide two working intervals for different fluid environments, further enhancing the underwater propulsion.

\section{Vehicle Design} \label{design}
\subsection{Design Methodology}
In comparison to vehicles that can operate exclusively in one medium, hybrid aerial-aquatic vehicles require additional consideration of the following concerns to improve their ability of operation in multiple mediums.
\begin{enumerate}
    \item \textbf{Cross-medium propulsion}: to generate enough lift in the air to overcome gravity and create sufficient propulsion in water. Additionally, optimize its propulsion efficiency in both mediums and lower overall weight.
    \item \textbf{Thruster configurations}: configure the rational arrangement of thrusters to overcome gravity and hydrodynamic resistance in the air and water, respectively.
    \item \textbf{Lightweight waterproofing}: while maintaining tightness and stability in water, lightening the waterproof construction to lower the workload on the aerial operation.
\end{enumerate}

To address the first issue, it is necessary to address the disparate demands placed on the operating point of the propulsion system by the vast disparity in density and viscosity of the two mediums. The magnitude of propulsion force is determined by the size of the mass flow created by the propulsion unit. For aerial conditions, fast sparse air flow should be provided by high speed propellers, whereas slow dense water flow should be generated by high torque driving propellers. To improve the performance of a cross-medium propulsion system, the motor and the propeller must be torque-matched in two different working ranges to accomplish the aforementioned output characteristics, which is challenging for a system that spans two mediums.

Hydrodynamics contain additional components: buoyancy, added mass, and hydrodynamic drag, necessitating consideration of the second issue in the design. For standard multirotor, thrusters are oriented vertically upwards to resist gravity, and yaw motion is generated by torque on propellers, which could be coupled with z-direction motion in the body frame and sluggish underwater (see Eq. (\ref{yaw_couple})). Surge or sway motion is accomplished by tilting the fuselage to direct small thrust portions in the horizon. Due to the presence of buoyancy and hydrodynamic drag, the fuselage must be tilted at a large angle to direct the majority of the thrust to the horizontal direction and accomplish the desired horizontal motion, which has been demonstrated in \cite{alzu2018loon}. The large-angle tilting has three drawbacks: firstly, its dynamic response underwater is obtuse due to added mass and hydrodynamic drag; secondly, the linear model based on the assumption of the small attitude angle fails; finally, it obstructs sensors, necessitating multiple sensor mounts. For underwater vehicles, thrusters are conventionally laid out in the direction of motion to translate with direct output and yaw with differential output, making it tough to achieve vehicle flight capabilities. Thus, the aerial-aquatic vehicle must have the aforementioned features to achieve efficient motion in both mediums.

In terms of waterproofing design, the sealability is determined by the joint surface smoothness and stiffness, and the underwater stability can be achieved by neutral buoyancy design and inherent restoring torque provided by the margin of the center of gravity and buoyancy. To achieve the lightweight requirements, solutions based on materials and structures have been investigated, such as rapid prototyping \cite{yu2020icuas}. The prototype in this study is designed to be lightweight at the expense of some sealability and can guarantee a maximum dive depth of 3m. 

Here, the waterproofing issue is not further discussed, and the multirotor-based design method is addressed below for the first two concerns.

\begin{figure}[tbp]
\centering
        \includegraphics[scale=0.45]{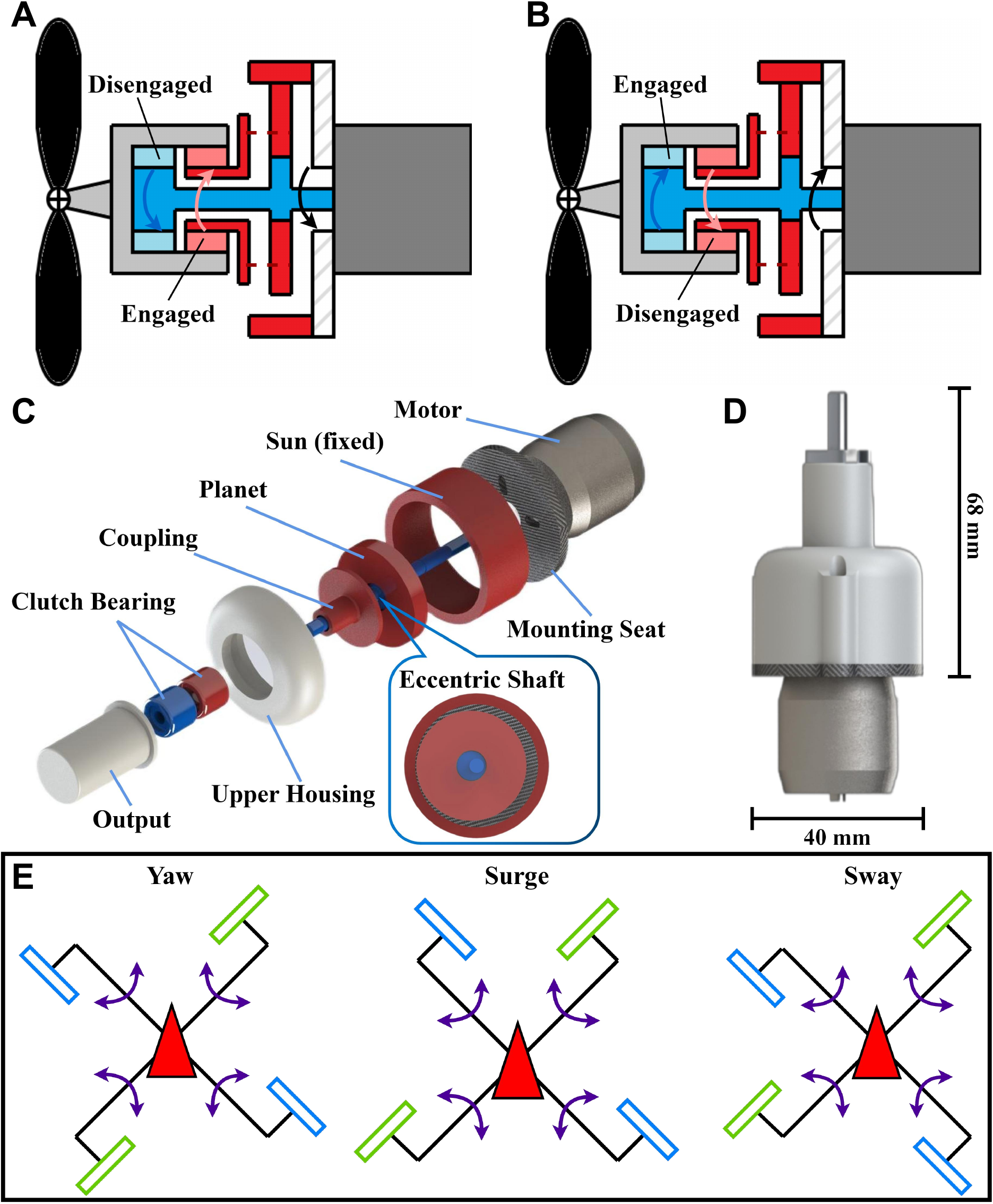}
\caption{A and B: The aquatic mode (A) and the aerial mode (B) are determined by the direction of the motor. C: The exploded view of the developed propulsion unit. D: The size of the propulsion unit. E: The schematic diagram of the tilt angle allocation.}\label{AAPU_combine}
\end{figure}

\subsection{Aerial-Aquatic Propulsion Unit}
There are three approaches to solving the torque matching problem for propulsion systems operating in two mediums. The first is installing two distinct sets of propulsion systems for water and air, which can provide excellent propulsion performance in both mediums but add an enormous workload \cite{bi2022nezha}. The second option is to employ one set of aerial propeller and motor to cope with both mediums \cite{li2022science}, the most straightforward and lightest approach. However, the issue is that improving the capacity to propel in one medium causes inevitable sacrificing on the ability to propel in the other. The final method uses one motor to drive two transmission chains for different medium environments and switches between them in time, including driving two propellers directly \cite{rockenbauer2021dipper} and driving one propeller through a gearbox at variable speed range \cite{yu2017ral}. This method is acceptable in weight and offers better propulsion in both water and air but has the disadvantage of a complex structure that could result in high operating losses and unreliability.

The developed prototype employs the above third solution of driving one propeller via a dual-speed gearbox. A less tooth differential planetary gear transmission is employed in the gearbox, which enhances reliability, and reduces the structure's complexity and the weight (the entire propulsion unit weighs 122g including the motor and the propeller). The schematic diagram of the working principle is presented in Fig. \ref{AAPU_combine}A and B. When the motor is in forwarding rotation, the thrust is transmitted by the blue chain and output directly for aerial mode; when the motor is in reverse rotation, the thrust is transmitted via the red one and output with greater torque for aquatic mode. This design has been validated in the prototype, and S. \ref{evaluation} will cover its performance evaluation.

\subsection{Independent Thrust Vectoring}\label{Vectoring Method}
Similar to the propulsion scheme, the thruster configuration problem has three solutions: first, equipped with two propulsion systems for different medium at the same time \cite{bi2022nezha}; second, directly use the air propulsion system underwater; and third, use a set of propulsion systems with variable structure and change the thruster configuration in due course.

Our idea is to modify the configuration by the thrust vectoring on one propulsion system, which has the advantage of higher integration and lighter weight of the entire system. In prior work, the coupled symmetric thrust vectoring mechanism provides the vehicle with dive mode and the angled mode which facilitates underwater movement. However, under the coupled symmetric method, it could be difficult to decouple thrusts generated by each propulsion and allocate them to a single degree of freedom due to the margin of the center of gravity and the action point of total thrust\cite{yu2020tmech}. To address this, each propulsion unit is modified to spin independently around its mount arm to vector thrusts. This design allows for the implementation of more sophisticated modes to investigate underwater mobility strategies based on independently thrust vectoring, as well as modes based on the original symmetric method.

\begin{figure}[tbp]
\centering
        \includegraphics[scale=2.4]{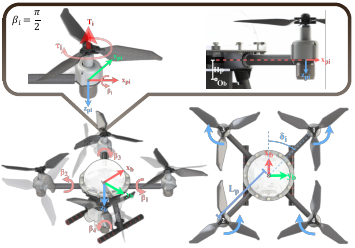}
\caption{Coordinate system definition: Body-fixed coordinate frame $\bm{F_b}$ and the $i$-th propulsion coordinate frame $\bm{{F_p}}_{i}$ associates with the thrust $T_i$, the torque $M_i$ and the tilting angle $\beta_i$. The symbol ${\rm H_p}$ and ${\rm L_p}$ represent the vertical and horizontal displacement from $\bm{F_b}$ to $\bm{{F_p}}_i$ respectively. The angle between $\mathbf{X_b}$ and the tilting arms is expressed as ${\rm \delta}_i$.}\label{overall}
\end{figure}

To analyze the effect of the propulsion unit tilting mechanism, the global North-East-Down (NED) coordinate frame introduced. The origin $\mathbf{O_b}$ of the body-fixed coordinate frame $\bm{F_b}$ is located at the center of gravity of the vehicle. And $\bm{{F_p}}_i$, $i = 1...4$, is defined as the coordinate system associated with the $i$-th propulsion unit group, with the origin attached on intersection of the tilting arm and the main shaft of the propulsion unit (see Fig. \ref{overall}). 

By denoting with $\mathbf{Z_{p}}_i$-axis and the origin $\mathbf{O_{p}}_i$ expressed in $\bm{F_b}$, the vectored propulsion component of i-th unit can be divided into the thrust part and the moment part as
\begin{align}
\small
    \begin{split}
        \mathbf{T}_i &= -T_i\mathbf{Z_{p}}_i \\
                     &= \begin{bmatrix}
                        -T_i c_{\beta_i}c_{\rm{\delta}_i}, &-T_i c_{\beta_i}s_{{\rm \delta}_i}, &T_i s_{\beta_i}
                        \end{bmatrix}^{\rm T},
    \end{split}\\
    \begin{split}
        \mathbf{M}_i &= bM_i\mathbf{Z_p}_i + \mathbf{O_{p}}_i \times(- T_i\mathbf{Z_{p}}_i)  \\
                     &= \begin{bmatrix}
                        bM_i c_{{\rm \delta}_i}c_{\beta_i} + T_i {\rm L_p} c_{{\rm \delta}_i}s_{\beta_i} - T_i {\rm H_p} s_{{\rm \delta}_i}c_{\beta_i}\\
                        bM_i s_{{\rm \delta}_i}c_{\beta_i} + T_i {\rm L_p} s_{{\rm \delta}_i}s_{\beta_i} + T_i {\rm H_p} c_{{\rm \delta}_i}c_{\beta_i}\\
                        - bM_i s_{\beta_i} + T_i {\rm L_p} c_{\beta_i}
                        \end{bmatrix},
    \end{split}
\end{align}where $b$ is a torque direction scalar which yields
\small{
\begin{equation}
    b = \left\{\begin{array}{cc}
         1  &,if \quad i=1,2 \\
         -1 &,if \quad i=3,4 
    \end{array}\right.,
\end{equation}
}and $s_*$, $c_*$, $t_*$ represents $sin(*)$, $cos(*)$, $tan(*)$.

Taking into account the introduction of the gearbox, the thrust $T_i$ and torque $M_i$ generated by the $i$-th propulsion unit is

\begin{align}
\small
    T_{i} &= \tilde{K}_{\rm T} \omega_{i}^2 = \frac{K_{\rm T}^{\star} \omega_{i}^2}{(r^{\rm \star})^2},\\
    M_{i} &= \tilde{K}_{\rm M} \omega_{i}^2 = \frac{K_{\rm M}^{\star} \omega_{i}^2}{(r^{\rm \star})^2},
\end{align}where $\omega_i$ is the rotational speed of the $i$-th motor, $K_{\rm T}^{\rm ae}$/$K_{\rm T}^{\rm aq}$ and $K_{\rm M}^{\rm ae}$/$K_{\rm M}^{\rm aq}$ are the aerial/aquatic constant coefficients of the propeller, $r^{\rm ae}$/$r^{\rm aq}$ are corresponding gear ratio, $\tilde{K}_{\rm T}$ and $\tilde{K}_{\rm M}$ are their calculated equivalent coefficients.

The mapping matrix from the motor speed $\mathbf{\Omega}$ to the twist $[T_{\rm z}, M_{\rm yaw}]^{\rm T}$ is introduced to analyze the previously mentioned issue of subpar underwater yaw motion performance of the conventional multirotor, as well as the benefits of the thrust vectoring:
\begin{equation}
\small
    \begin{bmatrix} T_{\rm z}\\ M_{\rm yaw} \end{bmatrix} = 
    \begin{bmatrix} \mathbf{K}_{\rm z}\\ \mathbf{K}_{\rm yaw} \end{bmatrix} 
    \mathbf{\Omega}\label{yaw_couple}.
\end{equation}

For quadrotor layout, $\mathbf{K}_{\rm z}=[\tilde{K}_{\rm T},\tilde{K}_{\rm T},\tilde{K}_{\rm T},\tilde{K}_{\rm T}]$, $\mathbf{K}_{\rm yaw}=[\tilde{K}_{\rm M},-\tilde{K}_{\rm M},\tilde{K}_{\rm M},-\tilde{K}_{\rm M}]$. Yaw movement with attitude level is achieved by differential output of motors: $\mathbf{\Omega}=[(\bar{\omega}+\Delta\omega)^2,\bar{\omega}^2,(\bar{\omega}+\Delta\omega)^2,\bar{\omega}^2]$. In this condition, the minimum output thrust is $T_z=2\tilde{K}_{\rm T}\Delta\omega^2$ with $\bar{\omega}=0$, which means that $T_{\rm z}$ and $M_{\rm yaw}$ can not be decoupled on a conventional quadrotor. In the air, the extra $T_{\rm z}$ can be used to balance gravity, but since the vehicle is designed to be approximately neutrally buoyant, the yaw motion could inevitably induce z-direction motion underwater.

For thrust vectoring, $\mathbf{K}_{\rm z}=\tilde{K}_{\rm T}s_{\beta_{i}}[1,1,1,1]$ and $\mathbf{K}_{\rm yaw}=(\tilde{K}_{\rm M}s_{\beta_{i}}+\tilde{K}_{\rm T}{\rm L_p}c_{\beta_{i}})[1,-1,1,-1]$. Yaw movement can be realized with $\beta_{i} = 0$ or $\pi$, which leads to $M_{yaw} = 4\tilde{K}_{\rm T}{\rm L_p}\bar{\omega}^2$ and $T_{\rm z} = 0$ with $\mathbf{\Omega}=[\bar{\omega}^2,\bar{\omega}^2,\bar{\omega}^2,\bar{\omega}^2]$. It means $T_{\rm z}$ and $M_{\rm yaw}$ can be decoupled, and since $\tilde{K}_T$ is dozens of times larger than $\tilde{K}_M$, the total yaw torque can be greatly increased. Thus, thrust vectoring allows for fast yaw motion during the underwater suspension, as demonstrated experimentally in S. \ref{Maneuver}.

A tilt angle allocation approach is designed to generate surge, sway, and yaw motions to demonstrate the ability of four separately driven thrust vectoring devices to function together. Fig. \ref{AAPU_combine}E illustrates such allocation, which can alternatively be expressed in the following mixer formula:

\begin{equation}
\small
    \begin{bmatrix} \beta_{\rm 1}\\ \beta_{\rm 2}\\ \beta_{\rm 3}\\ \beta_{\rm 4} \end{bmatrix} = 
    \frac{a\pi}{2}\Biggl(S\Bigl(\begin{bmatrix}
    -1 &  1 & -1\\
     1 & -1 & -1\\
     1 &  1 & -1\\
    -1 & -1 & -1
    \end{bmatrix}
    \begin{bmatrix} J_{\rm surge}\\ J_{\rm sway}\\ J_{\rm yaw} \end{bmatrix}\Bigr) - \begin{bmatrix} 1\\ 1\\ 1 \\1 \end{bmatrix}\Biggr),\label{mix}
\end{equation} where $S(x)$ is the sigmoid function that, combined with $a=1$ or $a=-1$, limits the angle to the range $(0, \pi)$ and $(-\pi, 0)$, portion of total the thrust vectoring operational range into two working zones, lower and upper, and $ J_* $  indicates the joystick volume for the surge, sway, and yaw channels.

The functionality of the approach is simply demonstrated by deploying the cascade PID controller of PX4 firmware without secondary development. Due to the robustness of the controller, the vehicle can remain stable under a certain tilt range of the propulsion unit, allowing horizontal and yaw motion without rotating the fuselage. S. \ref{Maneuver} shows the final implementation result and the comparison of maneuverability with the conventional quadrotor. The maneuverability is limited by the incompatibility of the controller, and the development at the control algorithm level needs to be further investigated.

\section{Prototype}\label{Proto}
\begin{figure}[thbp]
\centering
        \includegraphics[scale=0.41]{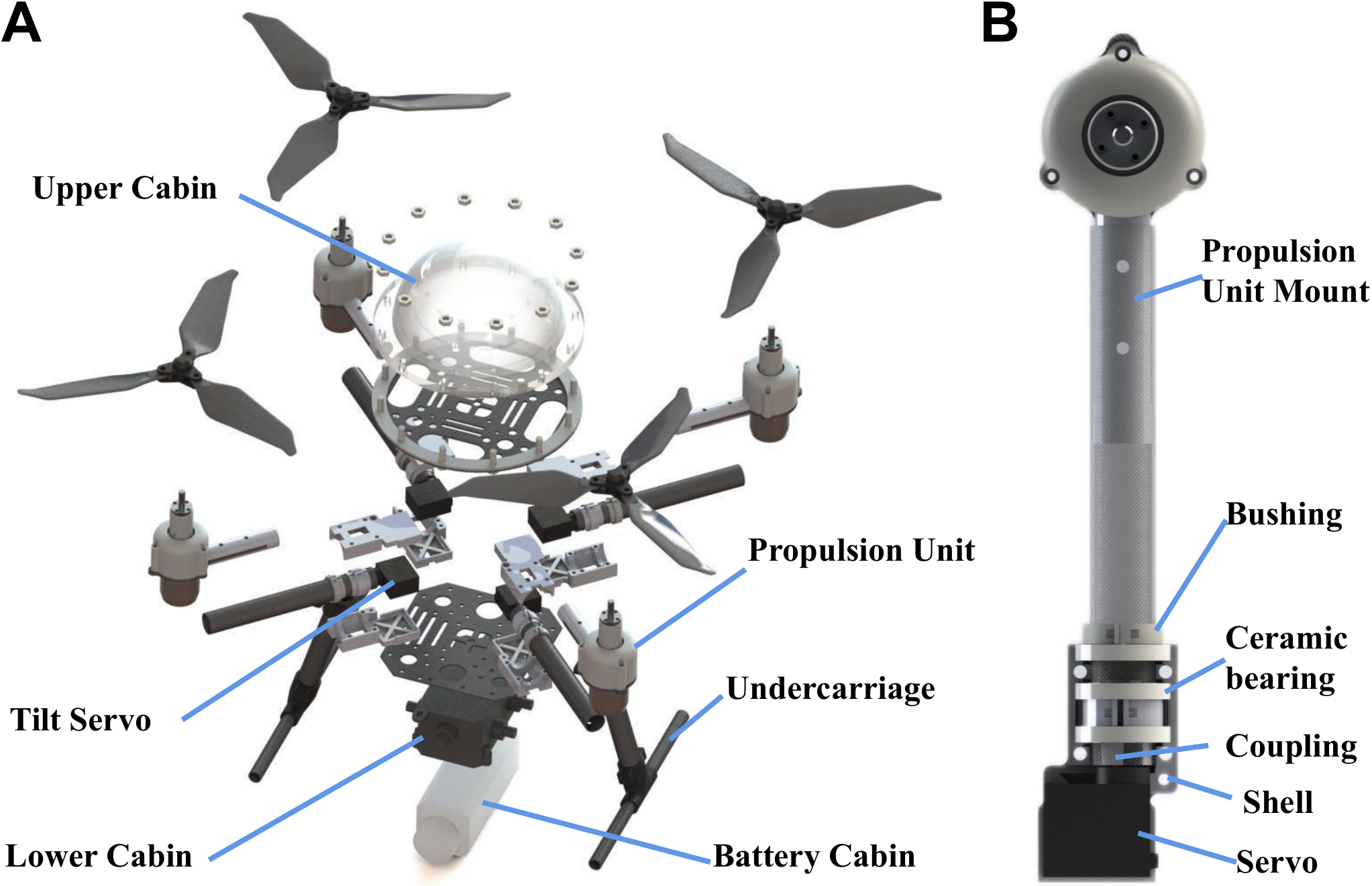}
\caption{A: The exploded view of the prototype. B: The tilt mechanism.}\label{Explode}
\end{figure}

A quadrotor prototype, TJ-FlyingFish, with independently tiltable propulsion units was built to verify the feasibility of the proposed concept in S. \ref{design}. It is with 380 mm wheelbase and 1.63 kg weight (aerial thrust to weight ratio: 3.75), can hover 6 minutes in the air and around 40 minutes underwater. It is composed of four major components: four independently operating arm tilting mechanisms located in the fuselage's center; four propulsion units installed at the ends of the arms; two watertight compartments located at the top and bottom of the tilting mechanisms to house the avionics system; and a watertight battery compartment located at the fuselage's bottom. Each component is depicted in exploded view in Fig. \ref{Explode}A. Additionally, it is designed to be under-buoyant in order to be suspended in water by the undeveloped control algorithm in PX4.

\subsection{Tilt Mechanism}
The similar mechanism exists in the aerial fully actuated multi-rotor \cite{ryll2014novel}, which is typically mounted at the end of the fixed arm and directly drives the rotor for tilting. Such design can not be fitted in the prototype due to the propulsion unit located at the end of the arm, as it could concentrate more mass at the end of the arm increases the rotational inertia and impairs the flight effect. Thus, the tilting mechanism is optimally mounted inside the body as illustrated in Fig. \ref{Explode}A.

However, the above design delivers the tilting force to the propulsion unit via the arm rather than directly, resulting in a somewhat long drive chain that, if structural clearance exists, could introduce the vibration to the entire system during operation and thus cause the wasteful energy loss and the controller collapse. To deal with this, three bearings are utilized to hold the tilt structure, as depicted in Fig. \ref{Explode}B, to ensure rigidity and eliminate the clearance with redundant support. The tilting arm is constructed of a carbon fiber tube with an internal coupling that connects to the servo motor, and bushings that attach to ceramic bearings. Finally, the parts mentioned above are inserted into shells and secured, and then forced together by the upper and lower carbon fiber covers that also contribute to attaching the waterproof cabins and other components.

\subsection{Component Selection}
\begin{figure}[htbp]
\centering
        \includegraphics[scale=0.96]{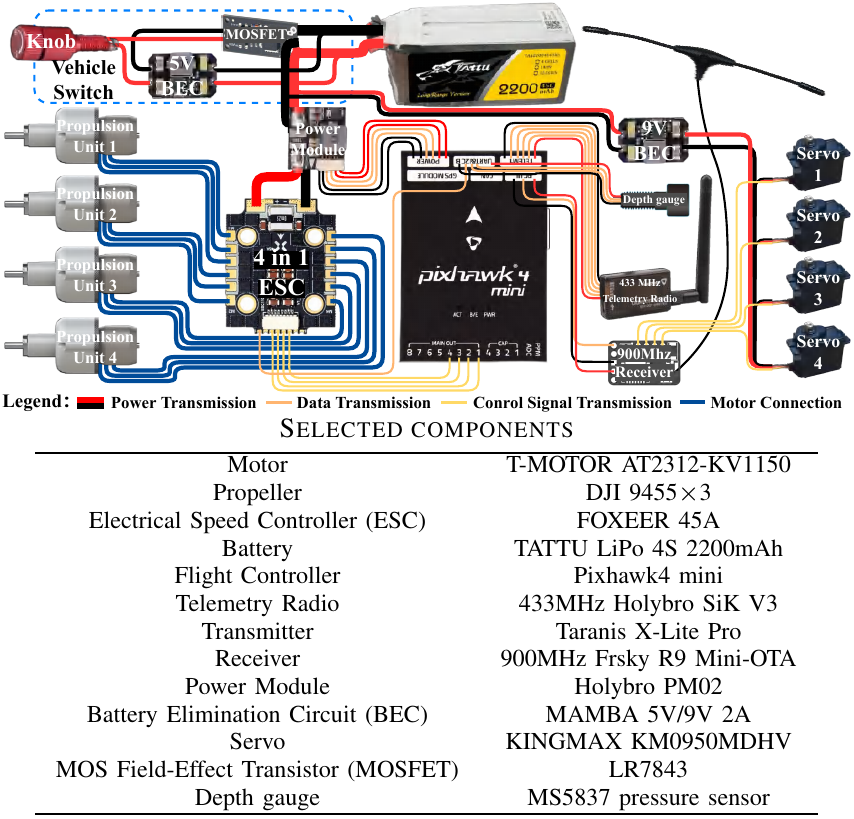}
\caption{The schematic diagram and the list of avionics system and actuators.}\label{Avionics}
\end{figure}

Components of the prototype and their wiring are described in Fig. \ref{Avionics}. The prototype is fitted with a 433MHz telemetry radio to monitor the prototype's status and alter parameters in real-time, which can be used routinely up to a depth of 2m underwater. A 900MHz radio system is utilized to assure the stability and the penetration of remote control signal underwater, which typically operates up to 1.5m deep. A 4-in-1 BLHeli32 ESC operates the four motors forward or backward by Dshot1200 signal and transmits the propulsion unit's real-time status to the Pixhawk4 mini. The real-time depth status is captured by an external integrated depth gauge. Additionally, a 9V BEC is used to power four high-voltage servos capable of producing 0.6N·m torque. Finally, a switching circuit operated by a MOSFET and a 5V BEC is equipped to avoid repetitive disassembly and assembly of the waterproof structure while powering on or off the vehicle.

\begin{figure}[thbp]
\centering
        \includegraphics[scale=0.95]{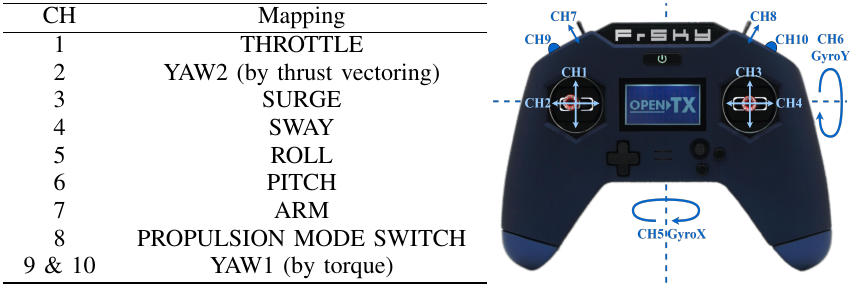}
\caption{The channel mapping on Frsky Taranis X-Lite Pro transmitter.}\label{Frsky}
\end{figure}

Due to the introduction of independent thrust vectoring, this prototype requires three more channels (SURGE, SWAY, and YAW2), in addition to the standard quadrotor's attitude control channels (ROLL, PITCH, and YAW1). The distinction between YAW1 and YAW2 is that YAW1 completes the yaw motion by torque, whereas YAW2 generates the yaw motion by thrust vectoring. As a typical transmitter would struggle to operate such many channels simultaneously, we picked the Frsky X-Lite Pro with an integrated gyroscope. The ROLL and PITCH channels are assigned to GyroX and GyroY, respectively, and YAW1 is mapped to the top buttons, allowing for the idle use of three joysticks to control the SURGE, SWAY, and YAW2 channels. Fig. \ref{Frsky} illustrates the exact remote control channel mappings.


Due to the introduction of the gearbox, the propeller and motor matching problem in the aquatic condition has been transformed into the transmission ratio design problem, and it concerns on the performance requirement of the aerial condition in the selection of motor and propeller. The T-MOTOR AT2312-KV1150 motor powered by the 4S LiPo battery and the three-blade carbon fiber propeller DJI 9455$\times$3 are optimally chosen for flight. According to the process mentioned in \cite{yu2017ral}, the ideal gear ratio for the above motor and propeller underwater is determined as 12.33 through QPROP simulation analysis. 

\section{Experimental Result}
The cross-medium propulsion unit and the independent thrust vectoring strategy are two contributing points of this work. To assess their improvement in aerial and aquatic locomotion, the propulsion ability is evaluated by the static output characteristics, and the aquatic maneuverability is by the comparison with the conventional multirotor.

\subsection{Static Output Characteristics of the Propulsion Unit}\label{evaluation}

\begin{figure}[tbp]
\centering
        \includegraphics[scale=0.49]{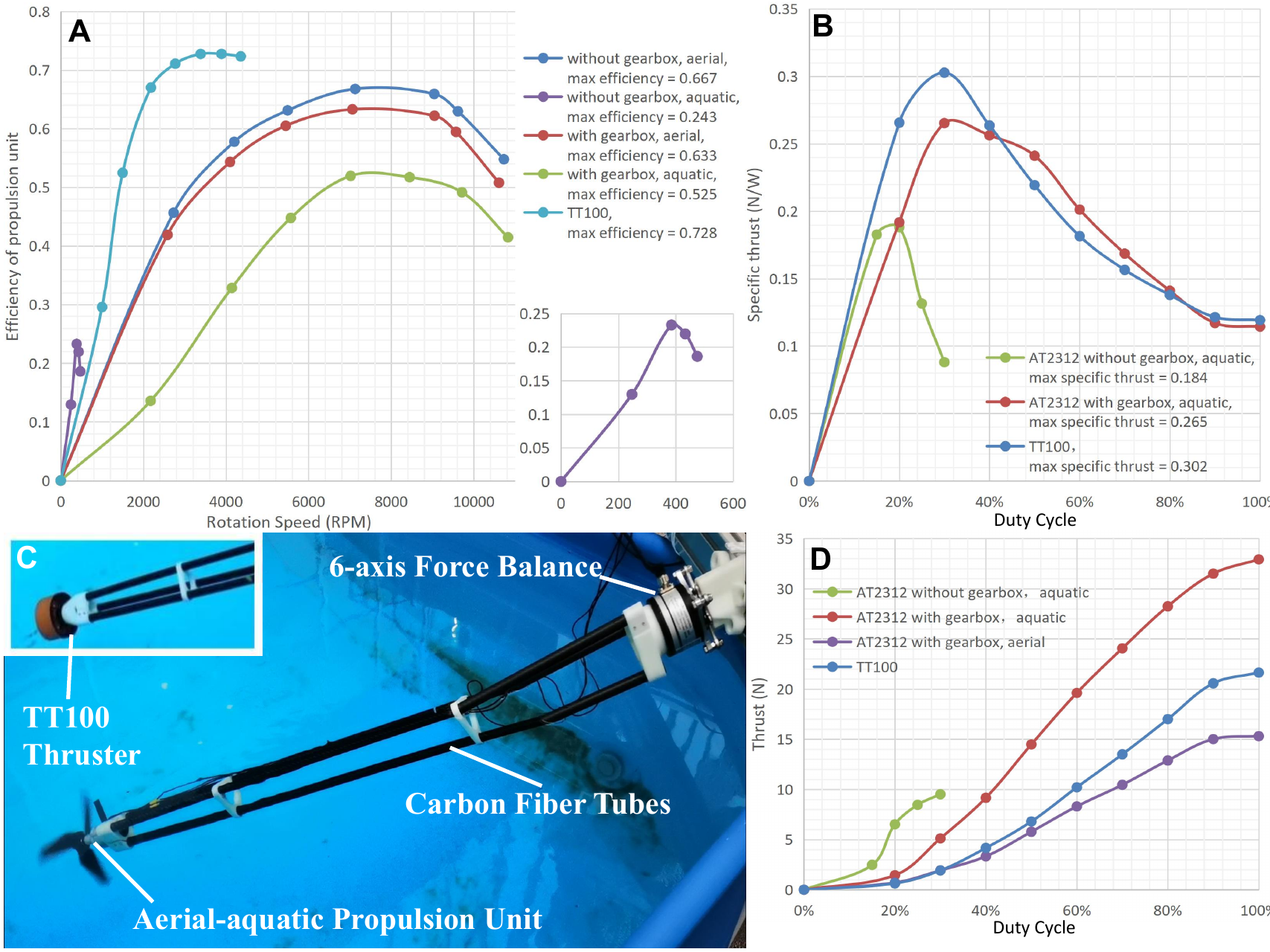}
\caption{ The static performance testing of the propulsion system: A: The efficiency of the motor and the gearbox. B and D: The specific thrust and thrust of the whole propulsion system. C: The propulsion test stand.}\label{performance}
\end{figure}

As shown in Fig. \ref{performance}C, the propulsion test stand is developed to examine the propulsion unit's actual performance. Due to non-waterproof of the 6-axis force balance (Maxwell F6D3934) used in the test, a cantilever beam structure constructed from three carbon fiber tubes is used to connect the propulsion unit to the balance, allowing force and moment output to be measured above the water surface. In addition, to compare the performance of the propulsion unit to that of a conventional underwater thruster, the TT100 thruster produced by TT robotix is tested as a comparator, which has a operating power nearly equivalent to that of the propulsion unit.

\begin{table}[tbp]
\fontsize{5.5pt}{8.0pt}\selectfont{
  \begin{center}
    \caption{Comparison of Aerial-Aquatic Propulsion Systems}\label{Mapping}
    \begin{tabular}{ccccc} 
      \bottomrule
      & \bf\tabincell{c}{Aerial\\Efficiency} 
      & \bf\tabincell{c}{Aquatic\\Efficiency} 
      & \bf\tabincell{c}{Max Aquatic\\Specific Thrust} 
      & \bf\tabincell{c}{Max Aquatic Thrust-\\to-Weight Ratio}\\
      \bottomrule
      \bf Proposed  & $63.3\%$ & $52.5\%$ & $0.265$ & $32N/122g=0.262$ \\
      \hline
      AquaMAV\cite{yu2017ral}             & $50.0\%$ & $46.0\%$ & $0.172$ & $9.5N/28g=0.339$ \\
      \hline
      MAAQuad\cite{yu2020icra}             & \tabincell{c}{$12.3\%$\\(simulated)} & unknown  & \tabincell{c}{$0.4$\\(simulated)}   &   \tabincell{c}{$4.8N/19.3g=0.249$\\(simulated)}\\
      \hline
      Loon-copter\cite{alzu2018loon}   & unknown & unknown   & $0.18$             & $12N/130.4g=0.092$              \\
      \hline
      Nezha-mini\cite{bi2022nezha}    & unknown & unknown   & $\approx0.27$ & $15N/83.4g=0.180$    \\
      \hline
      TT100         &  -      & $72.8\%$  & $0.302$       & $21.6N/290g=0.075$   \\
      \hline
    \end{tabular}
  \end{center}
  }\label{Comparison}
\end{table}

Fig. \ref{performance}A, B, and D illustrates the static output characteristics of them. Efficiency of the motor together with the gearbox is defined by the ratio of shaft power (shaft torque, $Q$ multiplied by angular speed, $\omega$) to electrical power (input voltage, $V$ multipied by input current, $I$):
\begin{equation}
\small
    \eta_{\rm M} = Q\omega/VI,
\end{equation}
and specific thrust is defined by the ratio of output thrust, $T$ to electrical power:
\begin{equation}
\small
    \Bar{T} = T/VI.
\end{equation}The aquatic efficiency $\eta_{\rm M}$ of the TT100 is higher than that of the propulsion unit (AT2312 with gearbox), while the specific thrust of this two is close. Because specific thrust measures the combined efficiency of the motor and the propeller, the above result indirectly indicates that the aerial propeller can also achieve high efficiency when employed underwater, which has been verified by simulation in \cite{yu2017ral}, and by experiment in \cite{alzu2018loon}. The presence of the gearbox unavoidably leads to the interal efficiency loss, and the testing data indicates a 5.1\% loss in the aerial operation, and a 21.3\% loss in the aquatic condition due to the open structure of the gearbox and the long transmission chain of aquatic mode, which introduces more hydrodynamic resistance and friction inside. Despite this, the gearbox structure still boosts the motor's operational efficiency as well as the total and specific thrust, and widens the operating range from 30\% to 100\% duty circle. Besides, its light weight of only 45g does not place an undue load on the air, ensuring aerial performance.

Table \uppercase\expandafter{\romannumeral1} shows a comparison of propulsion systems equipped in other aerial-aquatic vehicles. A similar concept is used on the AquaMAV \cite{yu2017ral}, which has a better thrust-to-weight ratio but poor static output characteristics. The MAAQuad \cite{yu2020icra} and the Loon-copter \cite{alzu2018loon} share the same work principle, propelled by aerial propulsion underwater, and only outperform the proposed in aerial characteristics. The Nezha-mini is propelled by two distinct systems that provide excellent performance in both mediums while lowering the thrust-to-weight ratio. Considering the efficiency, weight and output thrust comprehensively, the proposed scheme has better performance.

\begin{figure}[tbp]
\centering
        \includegraphics[scale=0.45]{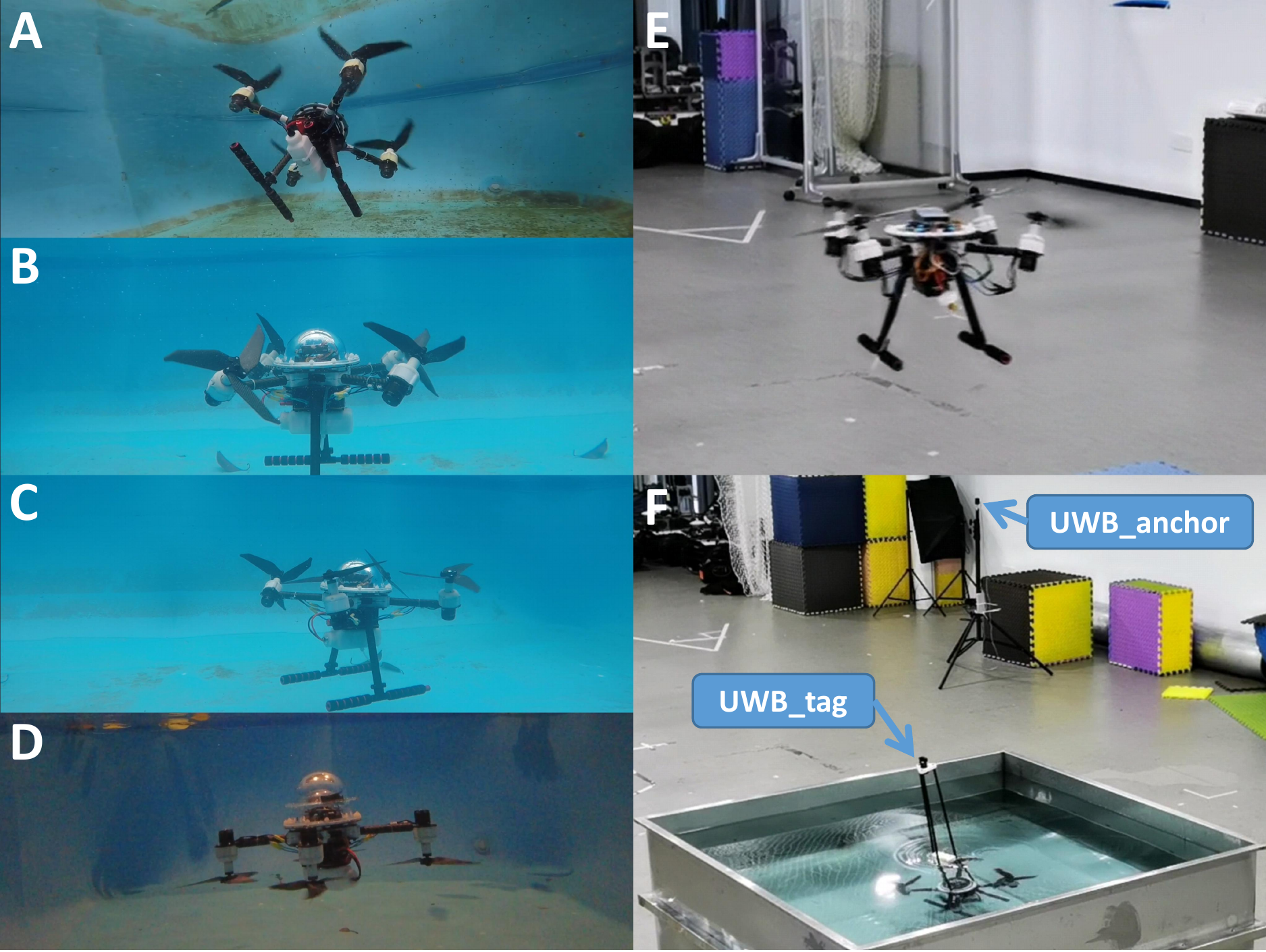}
\caption{Testing of the prototype: A: Horizontal move underwater by tilting the fuselage. B: Horizontal move underwater by thrust vectoring. C: Yaw underwater by thrust vectoring. D: Underwater dive mode ($\beta_i = -\frac{\pi}{2}$) testing with the additional buoyant block. E: Horizontal move in the air by thrust vectoring. F: Underwater positioning via UWB.}\label{vectoring_combine}
\end{figure}

\subsection{Aquatic Maneuverability} \label{Maneuver}

Fig. \ref{vectoring_combine}B-E displays snapshots of the independent thrust vectoring approach stated in S. \ref{Vectoring Method}. Although this approach is designed for underwater usage, experiments demonstrate that it can also be used in the air (Fig. \ref{vectoring_combine}E). Compared to the typical quadrotor maneuvering strategy (Fig. \ref{vectoring_combine}A), the thrust vectoring strategy significantly improves maneuverability, as evidenced in the yaw and horizontal movement operations. To evaluate underwater horizontal motion performance, UWB (Nooploop Linktrack S) is used for underwater velocimetry (Fig. \ref{vectoring_combine}F): long carbon fiber tubes connect the airframe to the UWB, which extends beyond water, and the underwater airframe position is solved by the attitude and the position of the UWB upon the surface.

\subsubsection{Underwater yaw movement}
\begin{figure}[thbp]
\centering
        \includegraphics[scale=0.90]{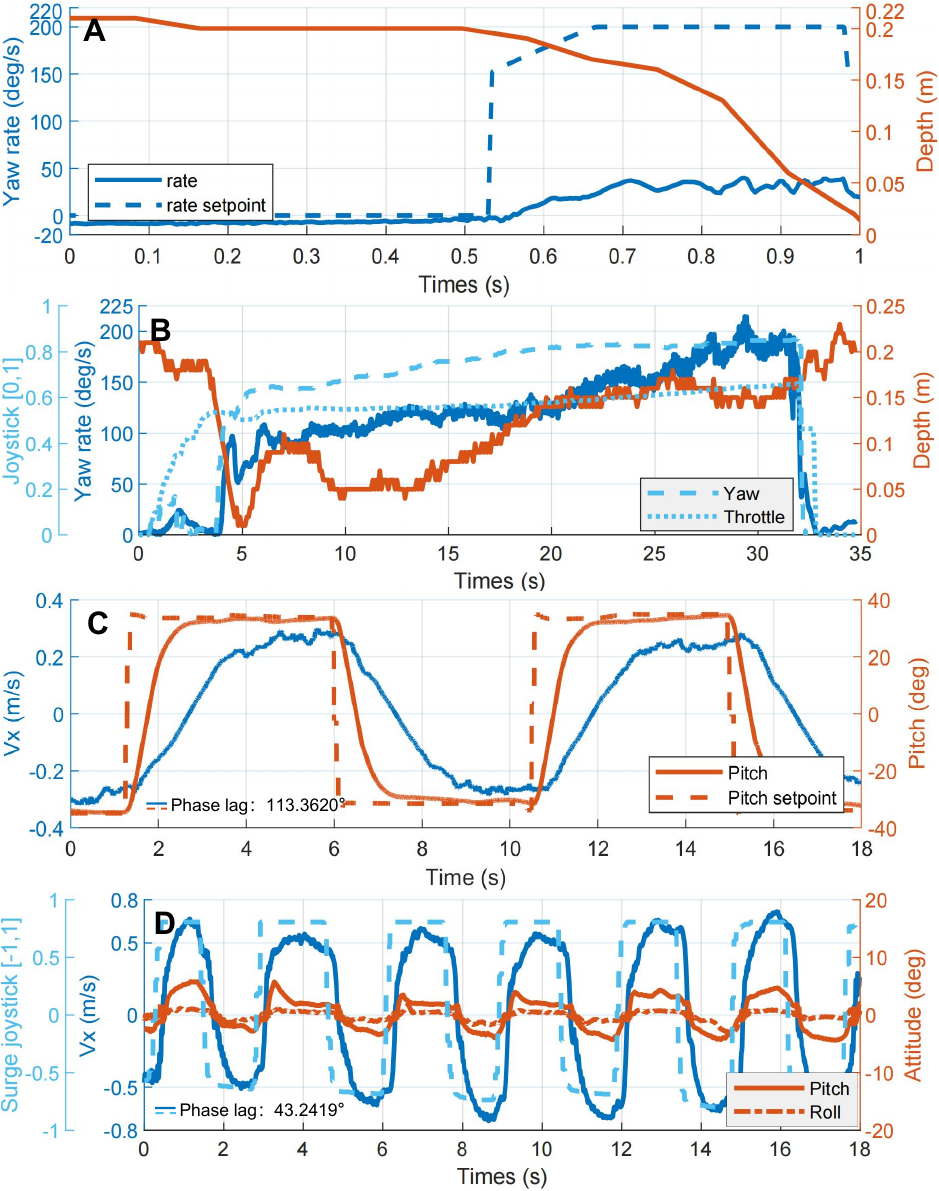}
\caption{Vehicle maneuverability testing: A: Yaw motion by torque. B: Yaw motion by tilt. C: Horizontal motion by tilting the fuselage. D: horizontal motion by thrust vectoring.  }\label{horizontalmove}
\end{figure}

Fig. \ref{horizontalmove}A and B shows the results of the implementation of yaw motion by torque (conventional quadrotor approach) and by thrust vectoring, respectively. Although the airframe is designed to be under-buoyant, when yawing by torque, the rate is still generated with the inevitable upward movement, and the rate magnitude is difficult to reach the setpoint value, confirming the problem of decoupling the yaw and the z-direction motion mentioned in S. \ref{design}. In the thrust vectoring process, the rotation angle of the arm is obtained by mapping the yaw joystick through Eq. (\ref{mix}), whose range is set to $[\frac{\pi}{6},\frac{5\pi}{6}]$ to generate the yaw moment and tiny thrust in the z-direction. The yaw motion with z-position holding (Fig. \ref{horizontalmove}B, $t = 20s\sim30s$) can thus be obtained by a coordinated manipulation of the throttle and yaw joystick, whose maximum rate is approximately 4-5 times than that by torque.

\subsubsection{Underwater horizontal movement}
For the horizontal motion by pitch (Fig. \ref{horizontalmove}C), the maximum body tilting angle is set to 35° (PX4 default), allowing the vehicle to travel horizontally while maintaining the z position at 0.3 m/s. For that by thrust vectoring (Fig. \ref{horizontalmove}D), the input of attitude is set to 0, and the arm rotation angle mapping in the range of $[\frac{\pi}{6},\frac{5\pi}{6}]$ is obtained by Eq. (\ref{mix}) in the same way as yaw movement, to provide horizontal thrust. The phase lag of $v_{\rm x}$ relative to the pitch setpoint in the former is around 113°, but the phase lag of $v_{\rm x}$ relative to the joystick in the latter is approximately 43°, indicating that the response speed of thrust vectoring underwater is greatly improved. However, although thrust vectoring achieved higher speeds in the experiment, this does not imply that the extreme speed is competitive. Due to the tilt arm mounting angle ${\rm \delta}_i$ (Fig. \ref{overall}), thrusts of each propulsion unit cannot be completely directed in the direction of motion, and partially cancel each other out, which is not the case while moving by tilting the fuselage, implying that increasing the tilt angle can result in a higher speed. Therefore, the benefit of thrust vectoring is not in fast underwater movement, but in the rapid and flexible adjustment of the direction of thrust to accomplish exact control of position and orientation.

The perturbation of attitude can be found in Fig. \ref{horizontalmove}D during thrust vectoring, which is due to two causes: first, there is no corresponding development of the controller for thrust vectoring, which also limits the experimental arm rotation angle in $[\frac{\pi}{6},\frac{5\pi}{6}]$ to prevent attitude controller from failure. Second, the dynamical coupling relation in the underwater environment is more complex than in the air, which is more severe for the non-streamlined fuselage, resulting in coupled attitude motion during horizontal motion. In response to the above two issues, further work will be done at the control algorithm level to improve the underwater operation performance.

\section{Conclusions}
This paper presents a quadrotor-based aerial-aquatic vehicle, TJ-FlyingFish, with tiltable dual-speed propulsion units. The practicality, functionality and maneuverability of the prototype have been thoroughly tested in the experiment. On one hand, the addition of the dual-speed gearbox enhances the aquatic propulsion characteristic while sacrificing little in terms of aerial performance. On the other hand, the implementation of independent propulsion vectoring mechanism enhances the aquatic agility of yaw and horizontal movement. Future work will be carried out at the level of  control algorithm, including the stability of the system under independent thrust vectoring and the corresponding control algorithm.


\balance
\bibliography{AAV}
\bibliographystyle{IEEEtran}

\end{document}